\pdfoutput=1

\documentclass[11pt]{article}

\usepackage[final]{acl}

\usepackage{times}
\usepackage{latexsym}
\usepackage{booktabs}
\usepackage{amsmath}

\usepackage{comment}
\usepackage{xspace}
\usepackage{booktabs}
\usepackage{graphicx}
\usepackage{amsmath}
\DeclareMathOperator*{\argmax}{argmax}
\usepackage{caption}
\usepackage{booktabs}  
\usepackage{adjustbox} 
\usepackage{float}
\usepackage[dvipsnames]{xcolor}
\usepackage{pgfplots}

\usepackage{pgfplotstable}
\usepackage{tikz}
\usepackage{array}
\newcolumntype{M}[1]{>{\centering\arraybackslash}m{#1}}

\pgfplotsset{compat=1.17}

\newcommand{\domenico}{}

\newcommand{\maru}{{Maru et al.\ 2022}\xspace}
\newcommand{\llamaone}{{Llama-3.2-1B-Instruct}\xspace}
\newcommand{\llamathree}{{Llama-3.2-3B-Instruct}\xspace}

\newcommand{\llamaeight}{{Llama-3.1-8B-Instruct}\xspace}
\newcommand{\llamaseventy}{{Llama-3.3-70B-Instruct}\xspace}

\newcommand{\gemmatwo}{{gemma-2-2b-it}\xspace}
\newcommand{\gemmafour}{{gemma-3-4b-it}\xspace}
\newcommand{\gemmanine}{{gemma-2-9b-it}\xspace}
\newcommand{\gemmatwelve}{{gemma-3-12b-it}\xspace}
\newcommand{\gemmatwentyseven}{{gemma-3-27b-it}\xspace}

\newcommand{\mistral}{{Mistral-7B-Instruct-v0.3}\xspace}
\newcommand{\ministral}{{Ministral-8B-Instruct-2410}\xspace}

\newcommand{\qwen}{{Qwen2.5-32B-Instruct}}

\newcommand{\phimini}{{Phi-3.5-mini-instruct}\xspace}
\newcommand{\phismall}{{Phi-3-small-128k-instruct}\xspace}

\newcommand{\deepseek}{{DeepSeek-V3}\xspace}
\newcommand{\gpt}{{GPT-4o}\xspace}

\usepackage[T1]{fontenc}

\usepackage[utf8]{inputenc}

\usepackage{microtype}

\usepackage{inconsolata}

\usepackage{graphicx}

\title{Do Large Language Models Understand Word Senses?}

\author{Domenico Meconi\textsuperscript{1} \quad Simone Stirpe\textsuperscript{1} \quad Federico Martelli\textsuperscript{2} \\ 
\quad \textbf{Leonardo Lavalle}\textsuperscript{2} \quad \textbf{Roberto Navigli}\textsuperscript{1,2} \\\\
\textsuperscript{1}Babelscape\\ 
    \textsuperscript{2}Sapienza NLP Group, Sapienza University of Rome \\
\texttt{\{surname\}@babelscape.com, \{surname\}@diag.uniroma1.it}
}

\begin{document}
\maketitle

\begin{abstract}

Understanding the meaning of words in context is a fundamental capability for Large Language Models (LLMs). Despite extensive evaluation efforts, the extent to which LLMs show evidence that they truly grasp word senses remains underexplored. In this paper, we address this gap by evaluating both i) the Word Sense Disambiguation (WSD) capabilities of instruction-tuned LLMs, comparing their performance to state-of-the-art systems specifically designed for the task, and ii) the ability of two top-performing open- and closed-source LLMs to understand word senses in three generative settings: definition generation, free-form explanation, and example generation. Notably, we find that, in the WSD task, leading models such as GPT-4o and DeepSeek-V3 achieve performance on par with specialized WSD systems, while also demonstrating greater robustness across domains and levels of difficulty. In the generation tasks, results reveal that LLMs can explain the meaning of words in context up to 98\% accuracy, with the highest performance observed in the free-form explanation task, which best aligns with their generative capabilities.
We release our code and data at: \url{https://github.com/Babelscape/LLM-WSD}.
\end{abstract}

\section{Introduction}

Thanks to the remarkable representational power of recent neural approaches based on the Transformer architecture \cite{vaswani2017attention} and its subsequent developments, language modeling has become the cornerstone of nearly every NLP application \cite{chang2023survey}: fueled by unprecedented scaling both in model size and training data, Large Language Models (LLMs) -- such as Llama \cite{touvron2023llama}, GPT \cite{brown2020language, achiam2023gpt}, and DeepSeek \cite{liu2024deepseek} among others -- have set new performance standards in a wide range of tasks.

While extensive evaluation has been conducted across numerous generative tasks,
a crucial yet underexplored capability of LLMs is their handling of lexical ambiguity, a pervasive feature of natural language \cite{bevilacqua2021recent}. Effectively resolving ambiguity is vital for generating contextually appropriate responses in NLP tasks such as translation and question answering, as misinterpretation can lead to hallucinations and reduced reliability \cite{Guerreiro:2023,campolungo2022dibimt}. Although many researchers believe that this issue is indirectly addressed by the current state-of-the-art performance across tasks, only a few studies have explicitly evaluated the models’ ability to disambiguate words in context. Importantly, recent research has revealed that LLMs struggle with ambiguous words, particularly when these are used in infrequent senses, either in translation tasks \cite{campolungo2022dibimt, martelli2025dibimt} or in other disambiguation-related evaluation scenarios \cite{basile2024ita, capone2024lost}. However,  no systematic study has yet addressed a fundamental research question: to what extent do LLMs understand the meaning of words in context?
With our work, we aim to bridge this gap with the following contributions:

\begin{itemize}
    \item We conduct an extensive experimental study on the explicit Word Sense Disambiguation (WSD) capabilities of open- and closed-weight LLMs, ranging in size from one billion to several hundred billion parameters. Our evaluation spans four distinct WSD benchmarks, including a newly curated dataset comprising over 5,500 items. We also compare the performance of the top LLM with that of an expert human annotator.    

    \item We investigate the lexical understanding abilities of the best-performing open- and closed-source models, i.e. \llamaseventy and \gpt, beyond the constraints of predefined sense inventories. Specifically, we evaluate them in three generation tasks: i) definition generation, ii) free-form explanation, iii) example generation. Our human-evaluated experiments reveal that both models demonstrate significantly enhanced understanding when allowed to define target words freely, achieving up to 98\% accuracy in the most unconstrained setting. 
\end{itemize}

\section{Related Work}

Over the course of the last few years, several benchmarks for LLMs have been proposed covering a wide range of language understanding tasks, such as the Massive Multitask Language Understanding benchmark \cite[MMLU]{hendrycks2020measuring}, Language Model Evaluation Harness \cite{eval-harness}, BIG-bench \cite{srivastava2023beyond}, InstructEval \cite{chia2023instructeval} and the Holistic Evaluation of Language Models \cite[HELM]{liang2023holistic}. However, despite their widespread adoption, such benchmarks do not specifically target disambiguation capabilities. 
Other studies have explored how LLMs handle various types of ambiguity, such as lexical, syntactic and pragmatic, via Natural Language Inference  \cite[NLI]{liu2023we, kamath2024scope}. Nevertheless, these works do not directly examine LLMs' ability to identify the correct sense of target words in context.

Among the research efforts which focus on this goal, \citet{kocon2023chatgpt} analyze the performance of ChatGPT 3.5\footnote{\url{https://openai.com/index/chatgpt/}} across various tasks, including  Word Sense Disambiguation (WSD) and Word-in-Context \cite{pilehvar2019wic}. However, their study is limited to a single model and does not offer a comprehensive view of LLMs' overall disambiguation capabilities.  
More recently, relatively low performance in terms of WSD capabilities has been reported by both \citet{yae2024leveraging} and \citet{kibria-etal-2024-functional}. However, while the former relies on only approximately 800 test instances, the latter does not adopt a robust evaluation strategy capable of handling unexpected LLM behaviors, such as hallucinations, inconsistencies, and deviations from the required response format. \citet{sumanathilaka2025llmsassistambiguityquantitative} explore a prompt augmentation strategy that incorporates additional linguistic cues such as part-of-speech tags, synonyms, and usage examples; their evaluation, nonetheless, is limited to a selected subset of the FEWS test set \cite{blevins-etal-2021-fews}.

Most closely related to our work, \citet{basile2024ita} propose a new benchmark in Italian for evaluating the disambiguation capabilities of LLMs, later extended to English by \citet{basile2025exploringwordsensedisambiguation}. However, both studies have limited model coverage and rely on automatically generated data such as XL-WSD \cite{pasinietal2021} and translated glosses derived from BabelNet \cite{navigli2012babelnet}.

While insightful, the aforementioned studies lack robust evaluation settings and do not offer comprehensive comparisons across LLMs spanning a wide range of model sizes -- from a few billion to hundreds of billions of parameters -- neither do they provide human-evaluated analyses of model performance. In this paper, we bridge this gap by first testing several instruction-tuned LLMs, i) on the WSD task, ii) in comparison with state-of-the-art WSD systems, iii) in zero-, one- and few-shot  scenarios. We then evaluate the best open- and closed-source LLMs in three different generation scenarios in which the models are free to express their ability to understand word senses.

\section{Lexical Understanding through WSD}
\label{sec:wsd}
We now investigate our first research question (\textbf{RQ1}): what is the capability of LLMs to carry out the definition selection task, i.e., Word Sense Disambiguation, where the model is provided with a word in context along with a list of dictionary definitions for the target word?

\begin{table*}[ht]
\centering
\small

\begin{tabular}{p{1cm}p{14cm}} 
\toprule

\textbf{Setting} & \textbf{Example Prompt} \\

\midrule

\textbf{WSD} & 
Read the sentence: \textit{I heard parts of the building above my head cracking.}  Choose the correct dictionary definition of the word ``\textit{building}" from the options: \\
& {1) A structure that has a roof and walls and stands more or less permanently in one place.} \\
& {2) The act of constructing something.}\\
& {3) The commercial activity involved in repairing old structures or constructing new ones.}\\
& {4) The occupants of a building.}\\
&Provide as output only the correct dictionary definition.\\
\bottomrule

\end{tabular}

\caption{Prompt example for WSD in a zero-shot scenario.}
\label{tab:example_prompts}

\end{table*}

\begin{table*}[ht]
\centering
\resizebox{\textwidth}{!}{
\begin{tabular}{lrrrrrrrrrrrrrrrrrrrr}

\cmidrule(lr){2-21}

& \multicolumn{20}{c}{\textbf{Prompt Templates}} \\ 

\cmidrule(lr){2-21}

& \multicolumn{5}{c}{\textit{instruction-based}} & \multicolumn{4}{c}{\textit{question-based}} & \multicolumn{3}{c}{\textit{conversational}} & \multicolumn{3}{c}{\textit{synonyms-based}} & \multicolumn{3}{c}{\textit{QA-based}} & \multicolumn{2}{c}{\textit{neg.-based}} \\ 


\toprule
\textbf{Model} & \textbf{P1} & \textbf{P2} & \textbf{P3} & \textbf{P4} & \textbf{P5} & \textbf{P6} & \textbf{P7} & \textbf{P8} & \textbf{P9} & \textbf{P10} & \textbf{P11} & \textbf{P12} & \textbf{P13} & \textbf{P14} & \textbf{P15} & \textbf{P16} & \textbf{P17} & \textbf{P18} & \textbf{P19} & \textbf{P20} \\
\midrule
\textbf{Llama-3.2-1B-it} & 36.7 & 50.3 & 35.4 & 45.7 & 33.9 & 52.5 & 40.9 & 39.1 & 44.4 & 51.2 & 46.8 & 45.5 & 33.6 & 38.2 & 33.2 & 45.1 & \textbf{51.6} & 46.6 & 36.7 & 31.2 \\
\textbf{Phi-3.5-mini-4B-it} & 58.5 & 58.2 & 58.9 & \textbf{60.0} & 58.0 & 55.8 & 57.1 & 58.0 & 59.6 & 54.3 & 56.9 & 56.3 & 55.2 & 53.8 & 53.2 & 56.5 & 55.9 & 56.3 & 57.6 & 55.8 \\
\cmidrule(lr){2-6} \cmidrule(lr){7-10} \cmidrule(lr){11-13} \cmidrule(lr){14-16} \cmidrule(lr){17-19} \cmidrule(lr){20-21}
\textbf{Phi-3-small-128k-7B-it} & 65.7 & 64.6 & \textbf{68.6} & 65.1 & 66.6 & 64.2 & 67.3 & 61.1 & 64.8 & 60.0 & 62.9 & 63.3 & 65.7 & 66.2 & 66.6 & 62.4 & 68.1 & 60.4 & 67.3 & 62.6 \\
\textbf{gemma-2-9b-it} & 67.5 & 67.5 & 66.6 & 66.2 & 65.5 & 65.7 & 65.9 & 64.4 & 66.8 & 61.8 & 63.5 & 64.0 & 63.5 & 64.8 & 63.3 & 65.5 & 67.3 & 62.0 & \textbf{68.6} & 67.3 \\
\cmidrule(lr){2-6} \cmidrule(lr){7-10} \cmidrule(lr){11-13} \cmidrule(lr){14-16} \cmidrule(lr){17-19} \cmidrule(lr){20-21}
\textbf{DeepSeek-V3} & 69.9 & \textbf{71.0} & 69.7 & 68.6 & 67.0 & 68.1 & 68.8 & 69.2 & 69.4 & 68.1 & 70.1 & 67.7 & 65.3 & 64.6 & 65.5 & 68.6 & 68.8 & 68.1 & 68.4 & 68.1 \\
\textbf{GPT-4o} & 71.6 & \textbf{72.1} & 69.9 & 70.1 & 71.9 & 69.7 & 70.1 & 71.6 & 70.8 & 71.2 & 70.1 & 72.1 & 68.6 & 71.2 & 70.8 & 71.4 & 71.2 & 70.6 & 71.6 & 66.8 \\
\cmidrule(lr){2-6} \cmidrule(lr){7-10} \cmidrule(lr){11-13} \cmidrule(lr){14-16} \cmidrule(lr){17-19} \cmidrule(lr){20-21}
\textbf{Mean} & 61.6 & \textbf{63.9} & 61.5 & 62.6 & 60.5 & 62.7 & 61.7 & 60.6 & 62.6 & 61.1 & 61.7 & 61.5 & 58.6 & 59.8 & 58.8 & 61.6 & 63.8 & 60.7 & 61.7 & 58.6 \\
\bottomrule
\end{tabular}
}
\caption{F1 scores obtained by representative models when evaluated in the WSD task on the SemEval-2007 dataset using 20  different prompts (see Appendix \ref{sec:prompts} for the list of prompts). Best result for each model in bold.}
\label{tab:dev-results}
\end{table*}

\subsection{Experimental setup}

\paragraph{Answer Extraction.}

Given a prompt containing a word in context and the list of candidate dictionary definitions for that word (see Table \ref{tab:example_prompts} for an example), we request the model to select the most appropriate definition among those provided and compute the highest lexical overlap score between the output definition and the candidate ones. Formally, let $\Delta_w$ be the set of candidate definitions associated with a given target word $w$ to be disambiguated, $\delta_\sigma$ the definition reproduced by the model and $\hat{\Delta}_w$ the set of gold definitions ($|\hat\Delta_w|\geq1$).  Subsequently, we tokenize\footnote{We use the NLTK library,  \url{https://www.nltk.org/}} both $\delta_\sigma$, resulting in $\delta_\sigma^{\tau}$ and all definitions contained in $\Delta_w$, thus obtaining $\Delta_w^{\tau}$. Then, we compute a lexical overlap score between $\delta_\sigma$ and all candidate definitions:

\[\delta_w^{max} = \argmax_{\delta_w \in \Delta_w} \frac{\left| \delta_\sigma^{\tau} \cap \delta_w^{\tau} \right|}{\left| \delta_w^{\tau}\right|}\]

Finally, if $\delta_w^{max} \in \hat{\Delta}_w$, we classify the answer as correct. Manual inspection of a substantial subset of model outputs confirms that the extracted answer consistently matches the model's intended selection in over 99.9\% of the cases on average.

We do not rely on perplexity to extract the model's answer, as we also evaluate closed-source models which do not provide access to logits. Nonetheless, we conduct some experiments using perplexity-based extraction on a subset of models (see appendix \ref{sec:perplexity}). Also, we do not utilize next-token probability, given our request to provide a full definition, and the well-known issues in the literature  \cite{wangetal:2024}.

\paragraph{LLMs.} We test instruction-tuned LLMs from the Gemma \cite{team2024gemma}, GPT \cite{hurst2024gpt}, DeepSeek \cite{liu2024deepseek}, Llama \cite{touvron2023llama}, Mistral \cite{jiang2023mistral}, Qwen \cite{bai2023qwen} and Phi \cite{li2023large} families. As \textbf{open-weight models}, we test \textbf{1-4B parameter models} (\llamaone, \llamathree, \gemmatwo, \gemmafour,  \phimini, \phismall ); \textbf{7-12B parameter models} (\mistral, \ministral, \llamaeight, \gemmanine, \gemmatwelve); \textbf{27-70B parameter models} (\gemmatwentyseven, \qwen, \llamaseventy); as \textbf{closed-source models}, we test two very large LLMs, namely, DeepSeek-V3 and GPT-4o. 
We test all the LLMs in zero-shot, one-shot and 3-shot settings (the latter providing examples with the same part of speech as the test item under consideration). More information about inference and hyperparameters can be found in Appendix \ref{sec:hyperparameters}.

\paragraph{Comparison models and baseline.} We test the LLMs against the highest-performance WSD systems, namely ConSec \cite{barba2021consec}, ESCHER \cite{barba2021esc} and BEM \cite{blevins-zettlemoyer-2020-moving}, all based on Transformer models.  
As our baseline, we compute the Most Frequent Sense (MFS), that  always selects as the answer the definition corresponding to the most frequent sense based on frequency in SemCor \cite{miller1993semantic}, as customary in the WSD literature. 

\paragraph{Test sets.} 
To evaluate the WSD capabilities of the models, we compute F1 performance on four different benchmarks, all annotated according to the WordNet sense inventory. Our primary test set is the English dataset proposed by \citet{maru2022nibbling}, a manually refined version of the popular WSD evaluation dataset by \citet{raganato2017word}, featuring $4917$ instances from the Senseval and SemEval competitions. 
We also include two supplementary datasets to test robustness across domains and difficulty levels, 42D and hardEN, also proposed by \citet{maru2022nibbling}. 42D contains $370$ instances sampled from 42 diverse domains, while hardEN is a subset of the primary test set, consisting of $476$ challenging cases on which all supervised WSD models failed. 
Finally, to address potential concerns about contamination from publicly available benchmarks -- which could compromise the validity of LLM evaluations -- we include a new WSD dataset explicitly curated for this study, hereafter referred to as WikiPortal WSD. It comprises 1,000 manually annotated sentences from news sources, totaling $5549$ instances, with a part-of-speech distribution that mirrors that of the \maru dataset. See Appendix \ref{sec:wikiportal} for further details.

\paragraph{Prompt design.} We design 20 distinct prompt templates that task the models to select the most appropriate definition for a given word in context. The  prompts vary not only in phrasing but also in the type of instruction they convey, enabling us to assess the sensitivity of model performance to different prompting styles. Specifically, we categorize the templates into six types: instruction-based, question-based, conversational, synonyms-based, question-answer-based and negation-based (see Appendix \ref{sec:prompts} for the full list). Each template includes a context sentence, the target word to be disambiguated, and the list of WordNet definitions for that word. We evaluate all prompts on the SemEval-2007 dataset \cite{agirre2007semeval}, which is typically used for development purposes. \domenico{To balance coverage and feasibility, we consider two representative LLMs from each parameter cluster introduced above (1-4B, 7-12B, 27-70B, and closed-source)}. As shown in Table \ref{tab:dev-results}, there is considerable variation in model performance across different prompt templates. The gap between the best and worst-performing prompts is about 5 points on average, with Prompt 2 achieving the highest overall mean score ($63.9$) and Prompts 13 and 20 the lowest ($58.6$). This variation highlights the sensitivity of LLMs (especially smaller ones) to prompt phrasing and structure in the context of lexical disambiguation. Given its top performance across models, we select Prompt 2 for further evaluation (an instance of which is shown in Table \ref{tab:example_prompts}).

\begin{table}[t]
\centering
\resizebox{1\linewidth}{!}{
\begin{tabular}{lccc}
\toprule
\textbf{Model} & \textbf{zero-shot} & \textbf{one-shot} & \textbf{few-shot} \\
\midrule
\textbf{Baseline} & $65.2$ & $65.2$ & $65.2$ \\
\midrule
\textbf{Llama-3.2-1B-it} & $56.3$ & $52.5$ & $60.3$ \\
\textbf{gemma-2-2b-it} & $72.4$ & $72.4$ & $73.0$ \\
\textbf{Llama-3.2-3B-it} & $68.4$ & $70.2$ & $68.3$ \\
\textbf{Phi-3.5-mini-4B-it} & $71.0$ & $72.3$ & $73.4$ \\
\textbf{gemma-3-4b-it } & \underline{$72.8$} & \underline{$74.0$} & \underline{$73.5$} \\
\midrule
\textbf{Phi-3-small-128k-7B-it} & $77.8$ & \underline{$80.1$} & $79.3$ \\
\textbf{Mistral-7B-it} & $74.3$ & $75.9$ & $75.3$ \\
\textbf{Ministral-8B-it} & $76.4$ & $75.8$ & $75.4$ \\
\textbf{Llama-3.1-8B-it} & $73.3$ & $73.9$ & $74.0$ \\
\textbf{gemma-2-9b-it} & \underline{$78.9$} & $78.4$ & \underline{$79.5$} \\
\textbf{gemma-3-12b-it} & $78.2$ & $78.0$ & $78.2$ \\
\midrule
\textbf{gemma-3-27b-it} & $78.8$ & $78.6$ & $78.8$ \\
\textbf{Qwen2.5-32B-it} & $78.8$ & $78.5$ & $79.5$ \\
\textbf{Llama-3.3-70B-it} & \underline{$81.4$} & \underline{$80.4$} & \underline{$81.2$} \\
\midrule
\textbf{DeepSeek-V3} & $81.9$ & $82.0$ & $82.0$ \\
\textbf{GPT-4o} & \textbf{82.3} & \textbf{82.6} & \textbf{83.2} \\
\midrule
\textbf{Mean} & $75.2$ & $75.4$ & $75.9$ \\
\bottomrule
\textbf{BEM} & $79.5$ & $79.5$ & $79.5$ \\
\textbf{ESCHER} & $81.6$ & $81.6$ & $81.6$ \\
\textbf{ConSeC} & \underline{$83.0$} & \underline{$83.0$} & \underline{$83.0$} \\
\bottomrule
\end{tabular}
}
\caption{F1 scores obtained by models when evaluated in the WSD task on the \maru test set. Best LLM in bold, best results for each cluster underlined.}
\label{tab:selection-summary}
\end{table}

\begin{table}[t]
\centering
\resizebox{0.75\linewidth}{!}{
\begin{tabular}{lcc}
\toprule
\textbf{Model} & \textbf{hardEN} & \textbf{42D} \\
\midrule
\textbf{Llama-3.2-1B-it} & $13.9$ & $24.3$ \\
\textbf{gemma-2-2b-it} & $31.3$ & $57.3$ \\
\textbf{Llama-3.2-3B-it} & $25.6$ & $53.5$ \\
\textbf{Phi-3.5-mini-4B-it} & \underline{$41.4$} & \underline{$66.2$} \\
\textbf{gemma-3-4b-it } & $34.5$ & $62.9$ \\
\midrule
\textbf{Phi-3-small-128k-7B-it} & $38.0$ & $67.0$ \\
\textbf{Mistral-7B-it} & $36.9$ & $63.5$ \\
\textbf{Ministral-8B-it} & $39.2$ & $67.6$ \\
\textbf{Llama-3.1-8B-it} & $31.5$ & $59.5$ \\
\textbf{gemma-2-9b-it} & \underline{$43.7$} & \underline{$70.0$} \\
\textbf{gemma-3-12b-it} & $37.4$ & $68.9$ \\
\midrule
\textbf{gemma-3-27b-it} & \underline{$44.1$} & $72.2$ \\
\textbf{Qwen2.5-32B-it} & $36.8$ & \underline{$72.7$} \\
\textbf{Llama-3.3-70B-it} & $37.8$ & $71.3$ \\
\midrule

\textbf{DeepSeek-V3} & $39.5$ & $74.6$\\
\textbf{GPT-4o} & \textbf{45.6} & \textbf{76.8} \\
\bottomrule
\textbf{BEM} & ~~$0.0$ & $53.2$ \\
\textbf{ESCHER} & ~~$0.0$ & \underline{$58.9$} \\
 \textbf{ConSeC} & ~~\underline{$7.4$} & $56.6$ \\
\bottomrule
\end{tabular}
}
\caption{F1 scores obtained by models when evaluated in the WSD task on the hardEN and 42D datasets in zero-shot settings. Best LLM in bold, best results for each cluster underlined.}
\label{tab:zero-shot-hard-42D}
\end{table}

\begin{table}[t]
\centering
\resizebox{1\linewidth}{!}{
\begin{tabular}{lccc}
\toprule
\textbf{Model} & \textbf{zero-shot} & \textbf{more-context} & \textbf{shuffle} \\
\midrule
\textbf{Baseline} & $65.2$ & $65.2$ & $65.2$ \\
\midrule
\textbf{Llama-3.2-1B-it} & $56.3$ & $56.2$ & $40.4$ \\
\textbf{gemma-2-2b-it} & $72.5$ & $72.4$ & $68.8$ \\
\textbf{Llama-3.2-3B-it} & $68.5$ & $68.2$ & $64.2$ \\
\textbf{Phi-3.5-mini-4B-it} & $71.0$ & $69.5$ & $73.3$ \\
\textbf{gemma-3-4b-it } & \underline{$72.9$} & \underline{$72.7$} & \underline{$70.3$} \\
\midrule
\textbf{Phi-3-small-128k-7B-it} & $77.8$ & $77.7$ & $76.3$ \\
\textbf{Mistral-7B-it} & $74.3$ & $74.2$ & $71.7$ \\
\textbf{Ministral-8B-it} & $76.5$ & $76.6$ & $73.9$ \\
\textbf{Llama-3.1-8B-it} & $73.3$ & $74.5$ & $70.4$ \\
\textbf{gemma-2-9b-it} & \underline{$78.9$} & \underline{$78.3$} & \underline{$78.1$} \\
\textbf{gemma-3-12b-it} & $78.2$ & $77.6$ & $77.5$ \\
\midrule
\textbf{gemma-3-27b-it} & $78.8$ & $78.5$ & $78.3$ \\
\textbf{Qwen2.5-32B-it} & $78.7$ & $78.6$ & $76.8$ \\
\textbf{Llama-3.3-70B-it} & \underline{$81.4$} & \underline{$81.8$} & \underline{$79.7$} \\
\midrule
\textbf{DeepSeek-V3} & $81.9$ & \textbf{82.9} & $80.3$ \\
\textbf{GPT-4o} & \textbf{82.3} & $81.7$ & \textbf{81.3} \\
\midrule
\textbf{Mean} & $76.5$ & $76.3$ & $74.7$ \\
\bottomrule
\end{tabular}
}
\caption{F1 scores obtained by models when evaluated in the WSD task on the \maru test set in three different settings: zero-shot, more-context, and shuffle. Best LLM in bold, best results per cluster underlined.}
\label{tab:zero-shot-context-shuffle}
\end{table}

\begin{table*}[t]
\centering
\small
\resizebox{1\linewidth}{!}{
\begin{tabular}{p{2.3cm}p{6.9cm}p{2.9cm}p{2.9cm}} 
\toprule
\textbf{Type} & \textbf{Sentence} & \textbf{Correct} & \textbf{GPT-4o} \\
\midrule
\textbf{Overgeneralization} & In the tower, five \textbf{men} and women pull rhythmically on ropes attached to the same five bells that first sounded here in 1614. & an adult person who is male (as opposed to a woman) & the generic use of the word to refer to any human being \\
\midrule
\textbf{Metonymy} & Ricardo Ulate said it's not surprising that the major powers are fighting over who should bear the costs for curbing greenhouse gases, even as vulnerable \textbf{countries} have become more aggressive in seeking to hold the big emitters accountable for their actions. & a politically organized body of people under a single government & the people who live in a nation or country \\
\midrule
\textbf{Gross\newline Misinterpretation} & The stranger, his head seemingly sunk in thought, \textbf{started} to cross the street against the light just as a huge moving van roared through the intersection.  & take the first step or steps in carrying out an action & move or jump suddenly, as if in surprise or alarm \\
\bottomrule
\end{tabular}
}
\caption{Examples of \gpt\ prediction errors.}
\label{tab:error-examples-selection}
\end{table*} 
\begin{table}[t]
\centering
\resizebox{1\linewidth}{!}{
\begin{tabular}{lccc}
\toprule
\textbf{Model} & \textbf{zero-shot} & \textbf{one-shot} & \textbf{few-shot} \\
\midrule
\textbf{gemma-3-4b-it } & $78.2$ & $77.8$ & $78.2$ \\
\textbf{gemma-2-9b-it} & $81.1$ & $81.3$ & $81.2$ \\
\textbf{Llama-3.3-70B-it} & $81.5$ & $81.2$ & $81.8$ \\
\textbf{GPT-4o} & \textbf{84.3} & \textbf{84.0} & \textbf{84.2} \\
\midrule
\textbf{ConSeC} & $82.2$ & $82.2$ & $82.2$ \\
\bottomrule
\end{tabular}
}
\caption{F1 scores obtained by models when evaluated in the WSD task on our new WikiPortal WSD dataset. Best results in bold.}
\label{tab:selection-new-dataset}
\end{table}

\subsection{Results}

\paragraph{Maru et al. 2022.}

We first report in Table \ref{tab:selection-summary} the results on the \maru WSD dataset (scores broken down by part of speech are available in Appendix \ref{sec:selection}). 
Overall, and as expected, among the evaluated LLMs, \gpt consistently achieves the highest F1 scores, followed by \deepseek, both of which are the largest models in the pool. Among the open-weight models, \gemmafour, \phismall, \gemmanine and \llamaseventy stand out as the leading options. The only model that does not surpass the MFS baseline ($65.2$) is the smallest one, \llamaone.

With a few exceptions, the impact of one and few-shot is negligible across all models, with the overall mean value (fourth-to-last row in the table) increasing from $75.2$ in the zero-shot setting to $75.9$ in the few-shot setting. For the best-performing model, i.e., GPT-4o, we observe a statistically significant improvement from zero to few-shot by almost one percentage point ($\chi^2$, $p < 0.05$). 

A key question is how LLMs compare to supervised systems explicitly trained for WSD. On this benchmark dataset, ConSeC and ESCHER attain $83.0$ and $81.6$, respectively, indicating  that \gpt and \deepseek are in the same ballpark, with no statistically significant difference compared to ConSeC ($\chi^2$, p < 0.05).

\paragraph{hardEN.} 
We now turn to the hardEN dataset (Table \ref{tab:zero-shot-hard-42D}, left column). Unsurprisingly, F1 scores drop significantly across all models with respect to the \maru benchmark. Even the strongest LLMs struggle on this challenging dataset: \gpt achieves the highest score at $45.6$, followed by \gemmatwentyseven at $44.1$. Among the smaller models, \gemmanine ($43.7$) and \phimini ($41.4$) stand out as the top performers, coming close to \gpt. It is worth noting that \gemmanine consistently performs well across both the \maru benchmark and hardEN, ranking among the top open-weight models in both settings. In contrast, \phimini performs moderately on the \maru benchmark, but shows competitive performance in hardEN. \domenico{While the performance of most LLMs is leveled around 40 F1 -- compared to 0 for most of the supervised WSD systems -- it is interesting to look at the complementary portion of the \maru benchmark: if we exclude the hardEN instances and consider only the remaining $4582$ items in \maru, ConSeC reaches an F1 score of $88.5$, outperforming \gpt, which scores $85.0$. Taken together, these results suggest that supervised systems like ConSeC excel on the "easier" cases but fail almost entirely on the hardest ones, whereas LLMs such as \gpt demonstrate a more balanced behavior across the board.}

\paragraph{42D.} 
We report results on 42D in the last column of Table \ref{tab:zero-shot-hard-42D}.  Compared to the other datasets, performance on 42D sits in the middle: scores are generally lower than on the \maru benchmark, yet higher than on hardEN. \gpt again leads with an F1 score of $76.8$, followed by \deepseek at $74.6$. Notably, \phimini and \gemmanine, which were already among the best-performing open-weight models on hardEN, continue to show good results, achieving $66.2$ and $70.0$, respectively. Interestingly, ConSeC and ESCHER show weaker performance compared to that achieved on the \maru benchmark. This highlights a key strength of LLMs -- namely, their wider coverage and better adaptability across domains.

\paragraph{Impact of context and candidate order.} 
We carry out two additional experiments in a zero-shot setting. The first investigates the effect of adding more context by extending the input prompt to include three sentences, where the target word appears in the middle one. The second experiment examines whether the order of candidate definitions influences model performance by randomly shuffling them. Examples of the prompt used in both experiments are provided in Table \ref{tab:example_prompts-context-shuffle}.

Table \ref{tab:zero-shot-context-shuffle} shows the impact of extended context (middle column) and shuffled definitions (last column). Interestingly, most models do not seem to benefit from the additional context. For instance, \gpt sees a slight, non-statistically significant decrease from $82.3$ to $81.7$. Notable exceptions include \deepseek, which improves from $81.9$ to $82.9$, and \llamaeight, which increases from $73.3$ to $74.5$ (both improvements are statistically significant, $p < 0.05$). More revealing are the results from the shuffled definition setting: when definitions are presented in random order, average performance across models drops from $76.5$ to $74.7$. The biggest drop occurs in the smallest model, \llamaone, showing a strong positional bias toward early options. This pattern highlights a potential limitation in some LLMs: rather than evaluating definitions purely on semantic appropriateness, they may rely on superficial heuristics such as positional order, which may compromise their reliability and generalization in unbiased WSD tasks.

\paragraph{Contamination check.} Table~\ref{tab:selection-new-dataset} reports the results on our newly introduced WikiPortal WSD dataset. We compute scores for the best-performing LLMs identified in our previous evaluations, along with the top supervised system, ConSeC. Consistently with the results on 42D and hardEN, \gpt achieves the highest F1 scores, reaching $84.3$ in zero-shot and outperforming ConSeC (a statistically significant difference). These results provide unequivocal evidence that high scores are not the result of test set contamination.

\paragraph{Systems vs. humans.}

Although top-performing LLMs approach the results of state-of-the-art WSD systems, they are still far from human performance. To quantify this gap, we conduct an experiment on a random subset of $400$ items from the \maru WSD test set,  where we ask an expert annotator with strong background in computational lexical semantics, paid according to the standards of their European country, to select the most appropriate sense for each instance. The annotator achieves an F1 score of $91.25$, compared to $82.5$ for \gpt on the same subset. This confirms that LLMs are not yet on par with human-level WSD performance (see also our evaluation of generation tasks in Section \ref{results_generation}).

\paragraph{Analysis.} To better understand the limitations of the best-performing model (i.e., \gpt) we manually analyze its incorrect predictions and identify six common error patterns. Table \ref{tab:error-examples-selection} reports three representative cases, while additional examples for the remaining patterns are provided in Appendix \ref{sec:errors}.
First, the model demonstrates \textbf{overgeneralization} errors, where it selects an overly broad or generic sense of a word. This is shown in the first row of Table \ref{tab:error-examples-selection}, where \gpt opts for a generic "human being" definition instead of "an adult person who is male".
Second, some errors arise from \textbf{metonymy}, where the model confuses closely related but distinct senses of a word. For example, for the word \textit{countries} \gpt selects the sense "a politically organized body of people under a single government" instead of the intended meaning "the people who live in a nation or country".
Finally, some errors are instances of \textbf{gross misinterpretation}, where the model selects a sense that clearly diverges from the context. As shown  in Table \ref{tab:error-examples-selection} (last row), the model interprets \textit{started} as "move or jump suddenly, as if in surprise or alarm", demonstrating a  misunderstanding of the context.

To validate the significance of these patterns and make a more meaningful comparison with humans (e.g. see the remarks by \citet{tedeschi-etal-2023-whats}), we manually categorize all 70 errors as either serious or borderline (the latter essentially due to the fine granularity of the sense inventory). 25 errors were tagged as \textbf{serious}, i.e., unlikely to be made by an expert human annotator. Interestingly, when we add these 25 to the 330 correctly classified items, the total (355) almost matches the performance of our human annotator in the same subset (88.75\% vs. 91.25\%). To cross-check these results, we also categorize the 35 errors made by our human annotator: only 2 were considered serious, due to oversight, while the remaining were marked as borderline cases (both validations were conducted  blindly by a second expert annotator). This corroborates the gap between GPT-4o (82.5\%) and the human annotator (91.25\%).
Moreover, only a third of the 45 borderline errors made by \gpt were also made by our annotator, suggesting that the ambiguity introduced by fine-grained sense distinctions leads to uncertainty, such that some items are correctly disambiguated by the model, and others by the human annotator.

\paragraph{Answer to RQ1.} \domenico{We find that top LLMs, such as \gpt and \deepseek, match or even surpass state-of-the-art supervised WSD systems like ConSeC and ESCHER. Overall, LLMs exhibit greater robustness across datasets compared to specialized WSD models, which are outperformed on hardEN and 42D by any 4B+ LLM. However, LLMs still lag behind humans and struggle with disambiguating hard and long-tail senses.}

\section{Lexical Understanding through Generation}
\label{results_generation}

\begin{table*}[ht]
\centering
\resizebox{1\linewidth}{!}{
\begin{tabular}{lcccccccccccc}

\cmidrule(lr){2-13}

& \multicolumn{12}{c}{\textbf{Dataset}} \\ 

\cmidrule(lr){2-13}

& \multicolumn{4}{c}{\textit{Maru et al. 2022}} & \multicolumn{4}{c}{\textit{42D}} & \multicolumn{4}{c}{\textit{WikiPortal WSD}} \\ 

\toprule
\textbf{Model} & \textbf{equally correct} & \textbf{better} & \textbf{worse} & \textbf{wrong} & \textbf{equally correct} & \textbf{better} & \textbf{worse} & \textbf{wrong} & \textbf{equally correct} & \textbf{better} & \textbf{worse} & \textbf{wrong}  \\
\cmidrule(lr){1-1}\cmidrule(lr){2-5} \cmidrule(lr){6-9} \cmidrule(lr){10-13}
\textbf{Llama-3.3-70B-it} & \multicolumn{1}{|c}{69.00} & 4.25 & 18.25  & 8.50 &  \multicolumn{1}{|c}{59.46} & 2.70 & 27.03 & 10.81 &  \multicolumn{1}{|c}{68.00} & ~~8.00 & 15.00 & 9.00 \\
\cmidrule(lr){2-13}
\textbf{GPT-4o} &  \multicolumn{1}{|c}{73.75} & 5.75 & 14.50 & 6.00 &  \multicolumn{1}{|c}{67.57} & 2.70 & 21.62 & ~~8.11 &  \multicolumn{1}{|c}{68.00} & 11.00 & 13.00 & 8.00 \\
\bottomrule
\end{tabular}
}
\caption{Results of \llamaseventy and \gpt on the Definition Generation task (\% of outputs tagged as equally correct, better, worse, or wrong) across three datasets.}
\label{tab:gen_eval_def}
\end{table*}
 
\begin{table*}[ht]
\centering
\resizebox{0.72\linewidth}{!}{
\begin{tabular}{lccccccccc}

\cmidrule(lr){2-10}

& \multicolumn{9}{c}{\textbf{Dataset}} \\ 

\cmidrule(lr){2-10}

& \multicolumn{3}{c}{\textit{Maru et al. 2022}} & \multicolumn{3}{c}{\textit{42D}} & \multicolumn{3}{c}{\textit{WikiPortal WSD}} \\ 

\toprule
\textbf{Model} & \textbf{good} & \textbf{bad} & \textbf{borderline} & \textbf{good} & \textbf{bad} & \textbf{borderline} & \textbf{good} & \textbf{bad} & \textbf{borderline}\\
\cmidrule(lr){1-1} \cmidrule(lr){2-4} \cmidrule(lr){5-7} \cmidrule(lr){8-10}
\textbf{Llama-3.3-70B-it} & \multicolumn{1}{|c}{95.75} & 2.50 & 1.75 & \multicolumn{1}{|c}{83.78} & 8.11 & 8.11 & \multicolumn{1}{|c}{98.00} & 2.00 & 0.00\\
\cmidrule(lr){2-10}
\textbf{GPT-4o} & \multicolumn{1}{|c}{96.00} & 3.00 & 1.00 & \multicolumn{1}{|c}{97.30} & 2.70 & 0.00 & \multicolumn{1}{|c}{97.00} & 3.00 & 1.00\\
\bottomrule
\end{tabular}
}
\caption{Results of \llamaseventy and \gpt on the Free-form Explanation task (\% of outputs tagged as good, bad or borderline) across three datasets.}
\label{tab:gen_eval_free}
\end{table*}
 
\begin{table*}[ht]
\centering
\resizebox{1\linewidth}{!}{
\begin{tabular}{lcccccccccccc}

\cmidrule(lr){2-13}

& \multicolumn{12}{c}{\textbf{Dataset}} \\ 

\cmidrule(lr){2-13}

& \multicolumn{4}{c}{\textit{Maru et al. 2022}} & \multicolumn{4}{c}{\textit{42D}} & \multicolumn{4}{c}{\textit{WikiPortal WSD}} \\ 

\toprule
\textbf{Model} & \textbf{yes} & \textbf{same pattern} & \textbf{no} & \textbf{borderline} & \textbf{yes} & \textbf{same pattern} & \textbf{no} & \textbf{borderline} & \textbf{yes} & \textbf{same pattern} & \textbf{no} & \textbf{borderline} \\
\cmidrule(lr){1-1}\cmidrule(lr){2-5} \cmidrule(lr){6-9} \cmidrule(lr){10-13}
\textbf{Llama-3.3-70B-it} & \multicolumn{1}{|c}{61.50} & 31.25 & 6.00 & 1.25 & \multicolumn{1}{|c}{70.27} & 10.81 & 16.22 & 2.70 & \multicolumn{1}{|c}{58.00} & 36.00 & 5.00 & 1.00 \\
\cmidrule(lr){2-13}
\textbf{GPT-4o} & \multicolumn{1}{|c}{59.75} & 33.50 & 4.50 & 2.25 & \multicolumn{1}{|c}{75.68} & 16.22 & ~~8.11 & 0.00 & \multicolumn{1}{|c}{57.00} & 36.00 & 6.00 & 1.00 \\
\bottomrule
\end{tabular}
}
\caption{Results of \llamaseventy and \gpt on the Example Generation task (\% of outputs tagged as yes, same pattern, no, or borderline) across three datasets.}
\label{tab:gen_eval_sent}
\end{table*}

\begin{table*}[t]
\centering
\small
\resizebox{1\linewidth}{!}{
\begin{tabular}{p{1.7cm}p{5cm}p{2.4cm}p{5.9cm}} 
\toprule
\textbf{Task} & \textbf{Context} & \textbf{Gold Definition} & \textbf{GPT-4o Generation} \\
\midrule
\textbf{Definition Generation} & Kagan served as solicitor general in the Obama administration when the first legal challenges to the \textbf{law} were brought at the trial court level. & legal document setting forth rules governing a particular kind of activity &  a system of rules created and enforced through social or governmental institutions to regulate behavior. \\
\midrule
\textbf{Free-form \newline Explanation} & In this sense, a \textbf{teacher} can be compared to one's family doctor. & a person whose occupation is teaching &  In the given sentence, the word "teacher" is being compared to a "family doctor," suggesting that a teacher, like a family doctor, plays a critical and foundational role in someone's life. 
\\
\midrule
\textbf{Example Generation} & \textbf{Ringers}, she added, are ""filled with the solemn intoxication that comes of intricate ritual faultlessly performed."" & a person who rings church bells (as for summoning the congregation) & The chess grandmasters sat in silence around the board, ringers fully immersed in the complexity and precision of their strategic moves, each one leading to a crescendo of intellectual triumph. \\
\bottomrule
\end{tabular}
}
\caption{Examples of \gpt\ generation errors.}
\label{tab:error-examples-generation}
\end{table*} 

While the above evaluation is an important step in positioning LLMs relative to state-of-the-art WSD systems in a classical setting, it reveals little about their ability to demonstrate an understanding of word senses in an unconstrained context, i.e., when they are free to answer without selecting from a predefined set of definitions, thus allowing them to express their full potential. This is the second research question of our paper (\textbf{RQ2}): to what extent can state-of-the-art LLMs explain the sense of words in context? Interestingly, this also helps address criticisms and limitations related to the discretization of sense distinctions in lexical semantics \cite{navigli2009word}. 

\subsection{Experimental Setup}

Previous attempts at evaluating the quality of generated definitions \cite{bevilacqua2020generationary} have relied on automatic metrics to assess the similarity between model outputs and gold definitions. In contrast, we aim to provide a more robust and reliable answer to the research question outlined above. \domenico{In order to do this, we randomly sample 550 items from three different datasets: the \maru WSD test set, 42D, and WikiPortal WSD. We task the two best-performing open-weight and closed-source models, \llamaseventy and \gpt, with answering three different types of prompts aimed at}: i) \textbf{definition generation}, ii) \textbf{free-form explanation}, iii) \textbf{example generation} (Table \ref{tab:example_prompts-generation} shows one example for each).
Then, we request an expert annotator with strong background in computational lexical semantics 
to tag the outputs generated by the models in the three tasks.

For the dictionary-style definition generation task (Table \ref{tab:error-examples-generation}, top) we ask the annotator to tag the definitions provided by the two LLMs in comparison with the WordNet definitions as selected in the gold standard dataset. For each item, the annotator has four options: \textit{equally correct, better than gold, worse than gold, wrong}. According to the guidelines, we consider a generated definition better if it more clearly captures the correct word sense, and worse if it is less appropriate or overly specific. \domenico{We count as evidence of correct sense understanding those outputs tagged as \textit{equally correct}, \textit{better than gold} or \textit{worse because overly specific}.}

In the free explanation task (Table \ref{tab:error-examples-generation}, center) the annotator has to evaluate the quality of the unconstrained explanation produced by the models for the target word in context. The evaluation options include \textit{good, bad} and \textit{borderline}. Borderline is used for those cases where the system grasps the intended meaning, but its output is either imprecise or overly tied to the example. \domenico{We consider that the model has correctly understood the target word's sense when its output is tagged as \textit{good}.}

In the example generation task (Table \ref{tab:error-examples-generation}, bottom), the annotator has to assess the ability of the LLMs to generate three distinct sentences that use the target word in the same sense as in the sentence under examination. In this case four options could be selected: \textit{yes} (all three sentences are accurate), \textit{yes but same pattern} (the sentences are correct but at least one contains a repetitive pattern that closely mirrors the input sentence), \textit{no} (at least one sentence is clearly incorrect), and \textit{borderline} (at least one sentence is ambiguous, leaving it open to multiple interpretations). \domenico{We consider that the model has correctly captured the intended sense when its output is tagged as \textit{yes} or \textit{yes but same pattern}.}
Full annotation guidelines and examples are available in Appendix \ref{sec:annotation}.

\subsection{Results} 

\paragraph{Definition generation.}
\domenico{Table 8 reports results for both \llamaseventy and \gpt across the three evaluation datasets. Overall, most of the generated dictionary-style definitions are judged to be of the same quality as the WordNet gold standard. Importantly, when models produce worse definitions, these are most often considered to be overly specific rather than incorrect or less appropriate. \gpt consistently outperforms \llamaseventy, with a gap of about 5\% in the \maru and WikiPortal WSD datasets. Specifically, in the \maru test set \gpt achieves 90.25\% satisfactory outputs (73.75\% equally correct, 5.75\% better, and 10.75\% overly specific), while \llamaseventy reaches 85\% (69.00\% + 4.25\% + 11.75\%). The gap increases to 8\% on 42D, where \gpt attains 81\% compared to 73\% for \llamaseventy, which we assume is due to the fact that \gpt is more robust across domains.}

\paragraph{Free-form explanation.}
\domenico{As shown in Table \ref{tab:gen_eval_free}, both \llamaseventy and \gpt achieve excellent results (up to 98\% answers rated as good) when let free to explain the word's sense, being in the same ballpark on \maru and WikiPortal (with no statistically significant differences). However, as in the definition generation task, a clear gap emerges on 42D, where \gpt outperforms \llamaseventy by 14\%.  These results show that placing LLMs in their most natural generative setting, i.e., completely free to explain, unlocks their full semantic potential.}

\paragraph{Example generation.}
\domenico{When tasked with the generation of same-sense usages of the word, the models exhibit mixed results (see Table \ref{tab:gen_eval_sent}). Across the three evaluation datasets, 60-70\% of the outputs successfully contained three distinct and semantically appropriate sentences. However, a significant portion (10-36\%) is tagged as relying on the same structural pattern as the input sentence, indicating a lack of syntactic variety while still preserving the intended word sense. As in the free-form explanation task, \llamaseventy and \gpt achieve similar performance on the \maru and WikiPortal WSD datasets, whereas on 42D \gpt outperforms \llamaseventy by 10\%. These results suggest that, while LLMs effectively capture the target word sense, they sometimes struggle to demonstrate flexibility in usage in various contexts.}

\paragraph{Inter-annotator agreement.} To corroborate the validity of our evaluation, we compute inter-annotator agreement with a second independent expert annotator on a subset of 100 items across all three generative tasks. We obtain Cohen’s $\kappa$ values of 0.791 for definition generation, 0.796 for free explanation, and 0.919 for example generation, indicating substantial to near-perfect agreement.

\paragraph{Analysis.} \domenico{As in Section \ref{sec:wsd}, we focus our error analysis on the type of errors made by \gpt across the three tasks.} In generating definitions, errors include gross hallucinations as in defining \textit{law} as "a system of rules created and enforced through social or governmental institutions to regulate behavior" in a context in which "legal document setting forth rules governing a particular kind of activity" is the right sense (Table \ref{tab:error-examples-generation}, top).

In the few errors observed in the free-form explanation task, \gpt tends not to provide an actual explanation, but rather to talk around the meaning, as in the example shown in Table \ref{tab:error-examples-generation}, middle, for the word \textit{teacher}.

When prompting the model to generate new examples, we still find gross hallucinations. For instance, considering the word \textit{ringer} (Table \ref{tab:error-examples-generation}, bottom), \gpt places the word in a context that does not align with its intended sense, failing to capture the correct sense.
In other cases, it fails to follow the prompt instructions, generating sentences where the target word is missing altogether.

\paragraph{Answer to RQ2.} \domenico{Removing the constraint of selecting from a predefined set of definitions allows state-of-the-art LLMs to explain the sense of words in context with considerably higher accuracy, above 90\% across the three generative tasks in an all-words scenario. LLMs reach up to 96-98\% in the free-form generation task, which aligns most naturally with their generative capabilities.}

\section{Conclusion}

In this paper, we present a comprehensive study on the ability of instruction-tuned Large Language Models to understand word meanings in context. We first evaluate a wide range of LLMs -- both open and closed-weight -- on four different  Word Sense Disambiguation (WSD) benchmarks, comparing their performance to state-of-the-art supervised systems. Top LLMs, such as \gpt and \deepseek, achieve performance comparable to specialized WSD models like ConSeC and ESCHER. Moreover, LLMs exhibit greater robustness on more challenging datasets, confirming their adaptability and reliability in the face of lexical ambiguity. Beyond classical disambiguation, we assess whether an open- or closed-source state-of-the-art model can demonstrate its semantic understanding in an unconstrained setting. Our analysis indicates that, regardless of the model or dataset, removing the constraints of predefined inventories allows the model to more accurately express the contextual meaning of words up to 98\% accuracy.

\section{Limitations}

All experiments were conducted exclusively in English. We leave the evaluation in different languages to future work. We note, however, that this goes beyond the scope of our paper, because for WSD very little data that is manually annotated exists in other languages. This pairs with the fact that  in languages other than English sense inventories are incomplete, which would make the results less robust (e.g. due to missing senses not given as additional potential options).

More LLMs could be tested, which we leave to future work. However, we posit that we evaluated a wide range of LLMs by parameter size and that the key findings in our paper are not negatively affected by this.

The generation component was evaluated only manually, which -- while avoiding well-known issues with the automated assessment of generated text -- may limit reproducibility. However, we release all the human answers for verification purposes, and in order to enable testing of future automatic evaluation approaches.

Finally, while our prompt design study explored a range of prompting strategies, we observed that performance varied across different formulations. This suggests that prompt sensitivity may play a more significant role than initially assumed, and future work could benefit from a deeper exploration of adaptive and task-specific prompting techniques.

\section*{Acknowledgements}

\begin{center}
\noindent
    \begin{minipage}{0.1\linewidth}
        \begin{center}
            \includegraphics[scale=0.05]{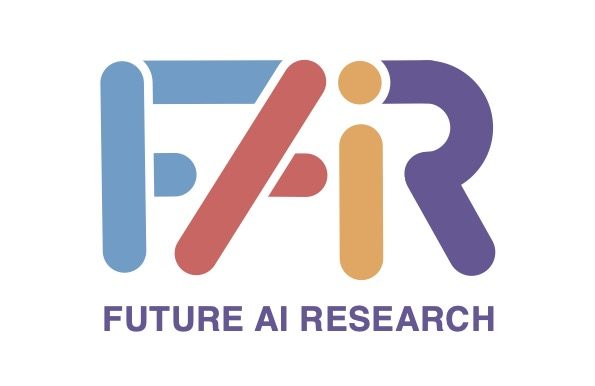}
        \end{center}
    \end{minipage}
    \hspace{0.01\linewidth}
    \begin{minipage}{0.70\linewidth}
         We gratefully acknowledge the support of the PNRR MUR project PE0000013-FAIR. 
    \end{minipage}
    \hspace{0.01\linewidth}
    \begin{minipage}{0.1\linewidth}
        \begin{center}
            \includegraphics[scale=0.08]{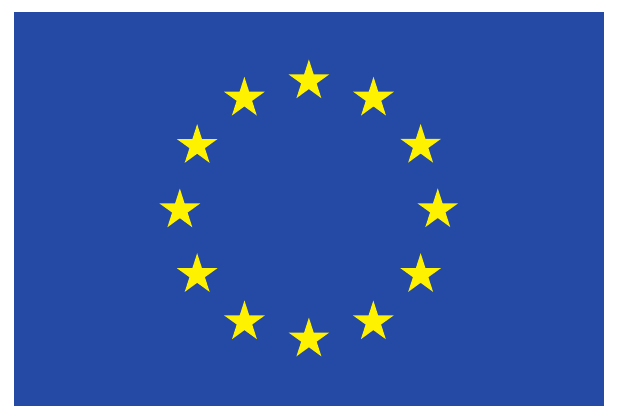}
        \end{center}
    \end{minipage}\\
\end{center}
\vspace{0.2cm}
\noindent We gratefully acknowledge the CREATIVE project (CRoss-modal understanding and gEnerATIon of Visual and tExtual content), which is funded by the MUR Progetti di Ricerca di Rilevante Interesse Nazionale programme (PRIN 2020).

\bibliography{custom}

\appendix

\section{Prompts}
\label{sec:prompts}

We report below the 20 prompt templates used in our evaluation of instruction-tuned LLMs on the Word Sense Disambiguation task. These prompts are organized into six categories, each reflecting a different prompting strategy or style. For ease of reference, each group of prompts is presented in a dedicated table: instruction-based (Table~\ref{tab:instruction-based}), question-based (Table~\ref{tab:question-based}), conversational-based (Table~\ref{tab:conversational-based}), synonyms-based (Table~\ref{tab:synonyms-based}), question-answer-based (Table~\ref{tab:qa-based}) and negation-based (Table~\ref{tab:negation-based}).

Table \ref{tab:example_prompts-context-shuffle} reports an example for more-context and shuffled-definitions experiments. 

Table \ref{tab:example_prompts-generation} shows an example for each of the three generation tasks.

\section{Perplexity}
\label{sec:perplexity}
In our evaluation, we primarily rely on lexical overlap between generated and gold definitions to extract model answers, as this method is straightforward and applicable to all models. However, to validate this choice, we performed additional experiments computing perplexity-based scores for a subset of open models where such access was possible.

In this approach, for each candidate definition, we compute the perplexity of the model generating that definition given the zero-shot prompt. The definition with the lowest perplexity is selected as the model's answer. This method captures the model's confidence in the candidate's fit to the context.

Table \ref{tab:perplexity} reports F1 scores for both extraction methods across several models in zero-shot setting. The results show that the scores based on lexical overlap and perplexity are very similar, with minimal differences for all models tested. This strong agreement supports the reliability of lexical overlap as a proxy for answer extraction.

\section{Implementation Details}
\label{sec:hyperparameters}
For smaller open-source models, all experiments were conducted on a single Nvidia 4090 GPU with 40 GB of RAM, using the \textit{transformers} library and its \textit{pipeline} method provided by HuggingFace\footnote{\url{https://huggingface.co/docs/transformers/index}}. 
For larger models, including those with 12B parameters or more, we relied on one node of a high-performance computing (HPC) cluster equipped with four A100 GPUs. In this case, we used the \texttt{vLLM} library\footnote{\url{https://docs.vllm.ai/en/latest/}} for efficient batched inference at scale.
For \gpt and \deepseek, inference was performed using the OpenAI API\footnote{\url{https://openai.com/index/openai-api/}}. 
We set the maximum number of output tokens to $75$ and left all other hyperparameters, such as temperature and top-$k$, at their default values.

Due to resource constraints, all experiments were run once without repeated trials.

\section{WikiPortal WSD Dataset}
\label{sec:wikiportal}
To mitigate potential test set contamination, we created a new Word Sense Disambiguation (WSD) dataset using 1000 manually selected sentences, totaling 5549 instances, sourced from the Wikipedia Current Events Portal\footnote{\url{https://en.wikipedia.org/wiki/Portal:Current_events}}. This portal aggregates recent news from a variety of international outlets, providing real-world, up-to-date content. The sources commonly used include Reuters, BBC News, Associated Press, Al Jazeera, The Guardian, and other international news organizations. Sentences were sampled from the last five years across domains and annotated with the appropriate sense using WordNet as the reference lexical resource. Annotations were carried out by an expert human annotator,  paid according to standards in their European country. We also confirmed the quality of annotations by asking a second annotator to sense tag a random subsample of 200 items, computing substantial $\kappa$ agreement, p < 0.05. \domenico{For confidentiality and to prevent potential test set contamination in future research, we will not make this dataset publicly available. However, we remain available to support validation of the dataset upon request.}

\section{WSD Results by Part of Speech}
\label{sec:selection}
Table~\ref{tab:zero-shot-general-results} reports the F1 scores obtained by all evaluated models on the WSD task on the \maru test set. Results are shown for three prompt settings: \textit{zero-shot}, \textit{one-shot}, and \textit{few-shot}, and are broken down by part of speech (NOUN, ADJ, VERB, ADV) as well as overall performance (ALL).

\section{Examples of \gpt WSD Errors}
\label{sec:errors}
Table \ref{tab:error-examples-selection-full} reports one representative example for each of the six error types identified in our analysis of the 70 incorrect predictions made by \gpt. Categories classified as serious errors, i.e., unlikely to be made by an expert human annotator, are: \textit{Gross Misinterpretation}, \textit{Literal vs. Metaphorical}, and \textit{Metonymy}.

\section{Human Annotation Guidelines and Examples}
\label{sec:annotation}
To provide transparency into the evaluation process, we report here the full annotation guidelines used for each of the three generative tasks: definition generation, free-form explanation, and example generation. 

\paragraph{Definition Generation.}
Annotators are evaluating the output of a computational linguistics experiment on the ability of LLMs to understand word meanings. 
For each instance, annotators are asked to assess the quality of a generated definition relative to a gold (reference) definition provided for a target word in context. The task consists in determining how well the generated definition captures the intended meaning of the word as it appears in its original usage. Annotators must choose one of four possible options: 

\textbf{Equal:} The generated definition is as appropriate as the gold definition. Both definitions capture the correct meaning of the word in context to a similar degree.

\textbf{Better:} The generated definition conveys the meaning of the word in context more clearly and accurately than the gold standard definition. It may also be more precise or informative.

\textbf{Worse:} The generated definition is less appropriate than the gold definition. It may be vague, overly specific, partially incorrect, or less clear in conveying the intended meaning of the word.

\textbf{Wrong:} The generated definition is incorrect, misleading, or not related to the meaning of the word in context.

Examples of annotations for each label are shown in Table \ref{tab:example-annotation-definition}.

\paragraph{Free-form Explanation.}
In this task, annotators are required to evaluate the quality of models' unconstrained explanation of a target word as used in context. For each item, the annotator must determine how well the explanation conveys the correct meaning of the target word in its specific context. The following three options are available: 

\textbf{Good:} This option must be selected when the explanation clearly and accurately conveys the correct meaning of the target word in context. 

\textbf{Bad:} This option must be selected when the explanation does not capture the intended meaning of the word in context, or when the model fails to provide a clear definition of the word used in the given sentence.

\textbf{Borderline:} This option is appropriate when the explanation demonstrates partial understanding: the model appears to grasp the correct sense of the word, but the output is imprecise, vague, or unsatisfactory.

Examples of annotations for each label are shown in Table \ref{tab:example-annotation-free}.

\paragraph{Example Generation.}
In this task, the annotator has to evaluate whether an LLM can generate three distinct example sentences that correctly use the target word with the same meaning it has in a provided reference sentence. The goal is to assess both the semantic accuracy of the examples and the variety of their syntactic or contextual formulations. The annotator must select one of the following four options:

\textbf{Yes:} This option has to be selected when all three generated sentences use the target word correctly, maintaining the same meaning as in the reference sentence.

\textbf{Yes but same pattern:} This option must be used when all three sentences are accurate in terms of word meaning, but follow a repetitive or overly similar pattern, closely resembling the structure or context of the original sentence. 

\textbf{No:} This option must be used when at least one of the three sentences clearly uses the word with a different meaning from that in the reference, or the usage is incorrect, misleading or contextually inappropriate.

\textbf{Borderline:} This option has to be selected when at least one sentence is ambiguous, meaning that the use of the target word could be interpreted as reflecting a different sense, even if the sentence is grammatically correct.

Examples of annotations for each label are shown in Table \ref{tab:example-annotation-example}.

\begin{table}[t]
\centering
\resizebox{1\linewidth}{!}{
\begin{tabular}{lcc}
\toprule
\textbf{Model} & \textbf{lexical-overlap} & \textbf{perplexity} \\
\midrule
\textbf{Llama-3.2-1B-it} & $56.3$ & $59.4$ \\
\textbf{gemma-2-2b-it} & $72.5$ & $72.4$ \\
\textbf{Llama-3.2-3B-it} & $68.5$ & $68.3$ \\
\textbf{Phi-3.5-mini-4B-it} & $71.0$ & $71.0$ \\
\textbf{gemma-3-4b-it } & $72.9$ & $72.7$ \\
\midrule
\textbf{Phi-3-small-128k-7B-it} & $77.8$ & $77.9$ \\
\textbf{Mistral-7B-it} & $74.3$ & $74.4$ \\
\textbf{Ministral-8B-it} & $76.5$ & $75.6$ \\
\textbf{Llama-3.1-8B-it} & $73.3$ & $74.9$ \\
\textbf{gemma-2-9b-it} & $78.9$ & $79.0$ \\
\bottomrule
\end{tabular}
}
\caption{F1 scores obtained by models when evaluated in the WSD task on the \maru test set using two answer extraction methods: lexical-overlap and perplexity}
\label{tab:perplexity}
\end{table}

\begin{table*}[ht]
\centering

\begin{tabular}{p{15cm}} 
\toprule

\textbf{Prompt template} \\

\midrule
\textbf{1)} Consider the following sentence: \textit{{<sentence>}} Select the most suitable dictionary definition which identifies the meaning of "\textit{{<word>}}" among the following definitions:\textit{{<definitions>}}. Provide as output only the most suitable dictionary definition. \\  
\midrule
\textbf{2)} Read the sentence: \textit{{<sentence>}} Choose the correct dictionary definition of the word “\textit{{<word>}}" from the options: \textit{{<definitions>}}. Provide as output only the correct dictionary definition. \\  
\midrule
\textbf{3)} In the sentence “\textit{{<sentence>}}”, determine which of the following definitions best represents the meaning of “\textit{{<word>}}": \textit{{<definitions>}}. Provide as output only the best definition. \\   
\midrule
\textbf{4)} Given the sentence: "\textit{{<sentence>}}", select the most appropriate meaning of the word “\textit{{<word>}}" from the choices: \textit{{<definitions>}}. Provide as output only the most appropriate definition. \\
\midrule
\textbf{5)} Consider the sentence: \textit{{<sentence>}} Identify the most accurate definition for “\textit{{<word>}}" from the following options: \textit{{<definitions>}}. Provide as output only the best definition.\\
\bottomrule
\end{tabular}
\caption{Prompt templates formulated as explicit instructions.}
\label{tab:instruction-based}

\end{table*}

\begin{table*}[ht]
\centering

\begin{tabular}{p{15cm}} 
\toprule

\textbf{Prompt template} \\

\midrule

\textbf{6)} What is the correct definition of "\textit{{<word>}}" in the sentence: "\textit{{<sentence>}}"? Choose from: \textit{{<definitions>}}. Provide as output only the correct definition. \\   
\midrule
\textbf{7)} In the sentence "\textit{{<sentence>}}", which of the following definitions best describes the meaning of "\textit{{<word>}}"? \textit{{<definitions>}}. Provide as output only the best definition. \\   
\midrule
\textbf{8)} Which of the following definitions correctly matches the meaning of "\textit{{<word>}}" in "\textit{{<sentence>}}"? \textit{{<definitions>}}. Provide as output only the best definition. \\   
\midrule
\textbf{9)} Based on the context of "\textit{{<sentence>}}", what does "\textit{{<word>}}" mean? Select from: \textit{{<definitions>}}. Provide as output only the best definition. \\  
\bottomrule
\end{tabular}
\caption{Prompt templates phrased as direct questions.}
\label{tab:question-based}

\end{table*}

\begin{table*}[ht]
\centering

\begin{tabular}{p{15cm}} 
\toprule

\textbf{Prompt template} \\
\midrule
\textbf{10)} If I ask you what "\textit{{<word>}}" means in "\textit{{<sentence>}}", how would you answer? Choose from: \textit{{<definitions>}}. Provide as output only the answer. \\   
\midrule
\textbf{11)} You are teaching vocabulary to a student. How would you explain the meaning of "\textit{{<word>}}" in "\textit{{<sentence>}}"? Select from: \textit{{<definitions>}}. Provide as output only the best definition. \\   
\midrule
\textbf{12)} Imagine you are explaining the meaning of "\textit{{<word>}}" in the sentence "\textit{{<sentence>}}", which definition would you choose? \textit{{<definitions>}}. Provide as output only the chosen definition. \\  
\bottomrule
\end{tabular}
\caption{Prompt templates designed in a conversational or instructional-dialogue style.}
\label{tab:conversational-based}

\end{table*}

\begin{table*}[ht]
\centering

\begin{tabular}{p{15cm}} 
\toprule

\textbf{Prompt template} \\
\midrule
\textbf{13)} In the sentence "\textit{{<sentence>}}", the word "\textit{{<word>}}" has a specific meaning. First, think about some synonyms that match its usage in this context. Then, choose the most appropriate definition from the following: \textit{{<definitions>}}. Provide as output only the best definition. \\   
\midrule
\textbf{14)} Consider the sentence \textit{{<sentence>}} Before selecting the correct meaning of "\textit{{<word>}}", think of a few synonyms that fit the context. Then, choose the best definition from: \textit{{<definitions>}}. Provide as output only the best definition. \\   
\midrule
\textbf{15)} Analyze the sentence \textit{{<sentence>}} Before determining the best definition for "\textit{{<word>}}", think of a few synonyms that could replace it in this context. Finally, choose the correct meaning from: \textit{{<definitions>}}. Provide as output only the best definition. \\
\bottomrule
\end{tabular}
\caption{Prompt templates that encourage the model to consider synonyms before selecting the most appropriate definition.}
\label{tab:synonyms-based}

\end{table*}

\begin{table*}[ht]
\centering

\begin{tabular}{p{15cm}} 
\toprule

\textbf{Prompt template} \\
\midrule
\textbf{16)} Question: What is the correct meaning of "\textit{{<word>}}" in the sentence: "\textit{{<sentence>}}"? Answers: \textit{{<definitions>}}. Provide as output only the correct definition. \\   
\midrule
\textbf{17)} Question: In the sentence "\textit{{<sentence>}}", which definition best represents the meaning of "\textit{{<word>}}"? Answers: \textit{{<definitions>}}. Provide as output only the best definition. \\   
\midrule
\textbf{18)} Question: Which of the following definitions correctly explains the meaning of "\textit{{<word>}}" in "\textit{{<sentence>}}"? Answers: \textit{{<definitions>}}. Provide as output only the best definition. \\
\bottomrule
\end{tabular}
\caption{Prompt templates structured in a labeled question–answer format.}
\label{tab:qa-based}

\end{table*}

\begin{table*}[ht]
\centering

\begin{tabular}{p{15cm}} 
\toprule

\textbf{Prompt template} \\
\midrule
\textbf{19)} In the sentence "\textit{{<sentence>}}", the word "\textit{{<word>}}" has a specific meaning. First filter out definitions that do not fit this context, and then select the correct one from the remaining choices: \textit{{<definitions>}}. Provide as output only the best definition. \\
\midrule
\textbf{20)} Consider the word “\textit{{<word>}}” in the sentence: \textit{{<sentence>}} Identify and exclude the incorrect definitions and then choose the best one: \textit{{<definitions>}}. Provide as output only the best definition. \\
\bottomrule
\end{tabular}
\caption{Prompt templates that ask the model to first eliminate inappropriate definitions before identifying the correct meaning of the word in context.}
\label{tab:negation-based}

\end{table*}

\begin{table*}[ht]
\centering

\begin{tabular}{p{1.5cm}p{13.5cm}} 
\toprule

\textbf{Setting} & \textbf{Example Prompt} \\

\midrule

\textbf{more-context} & 
Read the sentence: \textit{I rode it out on the second floor of Leo's at 55th and Telegraph in Oakland. I heard parts of the building above my head cracking. I actually thought that I might die.}  Choose the correct dictionary definition of the word ``\textit{building}" from the options: \\
& {1) A structure that has a roof and walls and stands more or less permanently in one place.} \\
& {2) The act of constructing something.}\\
& {3) The commercial activity involved in repairing old structures or constructing new ones.}\\
& {4) The occupants of a building.}\\
\midrule

\textbf{shuffled-definitions} & 
Read the sentence: \textit{I heard parts of the building above my head cracking.}  Choose the correct dictionary definition of the word ``\textit{building}" from the options: \\

& {1) The commercial activity involved in repairing old structures or constructing new ones.}\\
& {2) A structure that has a roof and walls and stands more or less permanently in one place.} \\
& {3) The act of constructing something.}\\
& {4) The occupants of a building.}\\
\bottomrule

\end{tabular}

\caption{Prompt examples (\textit{zero-shot}) for the WSD task with more context and shuffled definitions.}
\label{tab:example_prompts-context-shuffle}

\end{table*}
\begin{table*}[ht]
\centering

\begin{tabular}{p{2cm}p{13cm}} 
\toprule

\textbf{Task} & \textbf{Example Prompt} \\

\midrule

\textbf{Definition Generation} & Consider the following sentence: \textit{they needed rugs to cover the bare floors}. Provide a dictionary-style definition which identifies the meaning of ``\textit{floor}" in the above sentence. Do not motivate your answer and do not refer to the sentence context.\\

\midrule
\textbf{Free\newline Explanation} & Explain in your own words the meaning of the word ``\textit{floor}" in the following sentence: \textit{they needed rugs to cover the bare floors}.\\

\midrule
\textbf{Example Generation} &Provide three diverse examples that show me your understanding of the word ``\textit{floor}" used in the same meaning as the following sentence: \textit{they needed rugs to cover the bare floors}. List only the sentences.\\

\bottomrule

\end{tabular}

\caption{Prompt example for the three generation tasks.}
\label{tab:example_prompts-generation}

\end{table*}
\begin{table*}[t]
\centering
\resizebox{\linewidth}{!}{

\begin{tabular}{lccccccccccccccc}
\cmidrule(lr){2-16}

& \multicolumn{15}{c}{\textbf{Definition Selection}} \\ 

\cmidrule(lr){2-16}

& \multicolumn{5}{c}{\textit{zero-shot}} & \multicolumn{5}{c}{\textit{one-shot}} & \multicolumn{5}{c}{\textit{few-shot}} \\ 

\cmidrule(lr){2-6} \cmidrule(lr){7-11} \cmidrule(lr){12-16}

\textbf{Model} & \textbf{NOUN} & \textbf{ADJ} & \textbf{VERB} & \textbf{ADV} & \textbf{ALL} & \textbf{NOUN} & \textbf{ADJ} & \textbf{VERB} & \textbf{ADV} & \textbf{ALL} & \textbf{NOUN} & \textbf{ADJ} & \textbf{VERB} & \textbf{ADV} & \textbf{ALL} \\

\midrule
\textbf{Llama-3.2-1B-Instruct} & $58.92$ & $65.97$ & $44.51$ & $53.06$ & $56.34$ & $56.08$ & $55.22$ & $44.25$ & $38.27$ & $52.53$ & $64.26$ & $69.70$ & $47.35$ & $43.88$ & $60.30$ \\


\textbf{gemma-2-2b-it} & $75.21$ & $79.10$ & $61.77$ & \underline{$70.92$} & $72.48$ & $75.45$ & \underline{$81.04$} & $59.82$ & \underline{$70.92$} & $72.44$ & \underline{$76.82$} & $81.79$ & $59.20$ & \underline{$67.86$} & $73.09$ \\

\textbf{Llama-3.2-3B-Instruct} & $72.78$ & $75.52$ & $55.58$ & $54.08$ & $68.46$ & $73.47$ & $80.60$ & $58.14$ & $55.61$ & $70.21$ & $72.37$ & $75.22$ & $56.46$ & $52.04$ & $68.29$ \\

\textbf{Phi-3.5-mini-instruct} & $75.25$ & $77.46$ & $57.79$ & $62.76$ & $71.04$ & $76.51$ & $80.00$ & $57.96$ & $66.33$ & $72.32$ & $76.41$ & \underline{$82.24$} & $61.59$ & $66.84$ & $73.42$ \\

\textbf{gemma-3-4b-it} & \underline{$76.21$} & \underline{$79.40$} & \underline{$62.04$} & $62.76$ & \underline{$72.85$} & \underline{$77.85$} & $80.15$ & \underline{$61.86$} & $67.35$ & \underline{$74.07$} & $76.65$ & $81.64$ & \underline{$61.68$} & $65.82$ & \underline{$73.46$} \\

\cmidrule(lr){1-1} \cmidrule(lr){2-6} \cmidrule(lr){7-11} \cmidrule(lr){12-16}

\textbf{Phi-3-small-128k-instruct} & $81.38$ & \underline{$85.22$} & $65.31$ & $69.90$ & $77.75$ & \underline{$82.88$} & \underline{$87.76$} & \underline{$69.91$} & $71.94$ & \underline{$80.13$} & \underline{$82.16$} & $86.57$ & $68.58$ & $73.47$ & $79.30$ \\

\textbf{Mistral-7B-Instruct-v0.3} & $76.96$ & $80.45$ & $64.87$ & $67.86$ & $74.29$ & $78.60$ & $83.58$ & $65.40$ & $70.41$ & $75.92$ & $77.47$ & $84.93$ & $64.42$ & $71.94$ & $75.27$ \\

\textbf{Ministral-8B-Instruct-2410} & $78.64$ & $84.33$ & $66.73$ & \underline{$72.96$} & $76.45$ & $78.12$ & $82.69$ & $65.75$ & $74.49$ & $75.76$ & $77.82$ & $82.39$ & $65.13$ & $74.49$ & $75.39$ \\

\textbf{Llama-3.1-8B-Instruct} & $75.73$ & $81.94$ & $62.04$ & $71.43$ & $73.26$ & $76.65$ & $83.43$ & $61.77$ & $68.88$ & $73.85$ & $76.99$ & $83.58$ & $60.80$ & $70.92$ & $73.93$ \\

\textbf{gemma-2-9b-it} & \underline{$81.55$} & $84.18$ & \underline{$70.35$} & $71.43$ & $\underline{78.93}$ & $80.90$ & $83.88$ & \underline{$69.91$} & $72.45$ & $78.44$ & $81.44$ & $85.82$ & \underline{$71.59$} & \underline{$75.00$} & $\underline{79.52}$ \\

\textbf{gemma-3-12b-it} & $81.21$ & $83.88$ & $68.32$ & $70.92$ & $78.20$ & $80.18$ & $85.37$ & $68.14$ & \underline{$76.02$} & $77.95$ & $80.55$ & \underline{$86.87$} & $67.70$ & $73.98$ & $78.20$ \\

\cmidrule(lr){1-1} \cmidrule(lr){2-6} \cmidrule(lr){7-11} \cmidrule(lr){12-16}

\textbf{gemma-3-27b-it} & $80.83$ & $86.72$ & $70.88$ & $67.35$ & $78.81$ & $81.10$ & \underline{$86.57$} & $69.47$ & $66.33$ & $78.58$ & $81.24$ & $84.93$ & $70.09$ & $71.94$ & $78.81$ \\

\textbf{Qwen2.5-32B-Instruct} & $81.65$ & $84.18$ & $70.53$ & $64.80$ & $78.77$ & $81.79$ & $82.09$ & $69.47$ & $69.39$ & $78.50$ & $82.78$ & $82.99$ & $70.44$ & $69.90$ & $79.46$ \\

\textbf{Llama-3.3-70B-Instruct} & \underline{$83.84$} & \underline{$87.31$} & \underline{$72.57$} & \underline{$76.02$} & \underline{$81.41$} & \underline{$83.02$} & $85.07$ & \underline{$71.95$} & \underline{$73.98$} & \underline{$80.39$} & \underline{$82.99$} & \underline{$86.42$} & \underline{$73.54$} & \underline{$80.61$} & \underline{$81.19$} \\

\cmidrule(lr){1-1} \cmidrule(lr){2-6} \cmidrule(lr){7-11} \cmidrule(lr){12-16}

\textbf{DeepSeek-V3} & $84.39$ & \textbf{87.76} & $72.83$ & \underline{$76.53$} & $81.88$ & $84.32$ & \textbf{87.61} & $73.27$ & \underline{$77.55$} & $81.96$ & $84.46$ & \textbf{86.72} & $73.45$ & \underline{$77.55$} & $81.96$ \\

\textbf{GPT-4o} & \textbf{85.59} & $84.78$ & \underline{$74.51$} & $69.90$ & \underline{$82.31$} & \textbf{85.66} & $85.82$ & \textbf{75.13} & $70.41$ & \underline{$82.65$} & \textbf{85.72} & $86.57$ & \textbf{76.19} & $73.47$ & \textbf{83.16} \\

\cmidrule(lr){1-1} \cmidrule(lr){2-6} \cmidrule(lr){7-11} \cmidrule(lr){12-16}

\textbf{ESCHER} & $84.18$ & $85.82$ & $72.92$ & $78.06$ & $81.57$ & $84.18$ & $85.82$ & $72.92$ & $78.06$ & $81.57$ & $84.18$ & $85.82$ & $72.92$ & $78.06$ & $81.57$\\
\textbf{ConSeC} & \underline{$85.41$} & \underline{$86.56$} & \textbf{75.13} & \textbf{80.61} & \textbf{83.01} & \underline{$85.41$} & \underline{$86.56$} & \textbf{75.13} & \textbf{80.61} & \textbf{83.01} & \underline{$85.41$} & \underline{$86.56$} & \underline{$75.13$} & \textbf{80.61} & \underline{83.01}\\
\bottomrule

\end{tabular}
}

\caption{F1 scores obtained by models, broken down by part of speech (POS), when evaluated on the WSD task on the \maru test set. The best result for each POS is shown in bold, and best result within each cluster is underlined.}
\label{tab:zero-shot-general-results}

\end{table*}

\begin{table*}[t]
\centering
\small
\resizebox{1\linewidth}{!}{
\begin{tabular}{p{2.8cm}p{6.2cm}p{3cm}p{3cm}} 
\toprule
\textbf{Type} & \textbf{Sentence} & \textbf{Correct} & \textbf{GPT-4o} \\
\midrule
\textbf{Overgeneralization} & When functioning normally, they make proteins that hold a cell's \textbf{growth} in check. & (biology) the process of an individual organism growing organically;  & a process of becoming larger or longer or more numerous or more important \\
\midrule
\textbf{Over specificity} & To the extent that Democratic legislators from the South have \textbf{held} a disproportionate share of power in Congress since 1932 and have been able to translate such clout into relatively more local benefits for their respective constituencies ... & have or possess, either in a concrete or an abstract sense  & have rightfully; of rights, titles, and offices \\
\midrule
\textbf{Borderline} & The damage brought about to the \textbf{principles} of the sport by these actions is not as much ethical as aesthetic.  & a basic truth or law or assumption & a rule or standard especially of good behavior \\
\midrule
\textbf{Gross\newline Misinterpretation} & If they put a Republican into office, not only will they acquire less in terms of local \textbf{benefits} but their selected legislator will be relatively powerless to prevent other legislators from `` bringing home the bacon `` to their respective constituencies.  & financial assistance in time of need & something that aids or promotes well-being  \\
\midrule
\textbf{Literal\newline vs.\newline Metaphorical} & He \textbf{theorized} that in the eye cancer, an infant inherited a damaged copy of a gene from one parent and a normal copy from the other. & to believe especially on uncertain or tentative grounds & form or construct theories \\
\midrule
\textbf{Metonymy} & Ricardo Ulate, a Costa Rican delegate, said it 's not surprising that the major powers are fighting over who should bear the costs for curbing greenhouse gases, even as vulnerable \textbf{countries} have become more aggressive in seeking to hold the big emitters accountable for their actions. & a politically organized body of people under a single government & the people who live in a nation or country \\
\bottomrule
\end{tabular}
}
\caption{Examples of \gpt\ prediction errors.}
\label{tab:error-examples-selection-full}
\end{table*} 
\begin{table*}[t]
\centering
\small
\resizebox{1\linewidth}{!}{
\begin{tabular}{p{2cm}p{6cm}p{3cm}p{4cm}} 
\toprule
\textbf{Annotation} & \textbf{Context} & \textbf{Gold Definition} & \textbf{GPT-4o Generation} \\
\midrule
\textbf{Equal} & In the tower , five \textbf{men} and women pull rhythmically on ropes attached to the same five bells that first sounded here in 1614 . & an adult person who is male (as opposed to a woman)  & man (noun): an adult male human being. \\
\midrule
\textbf{Better} & I was just sitting down to meet with some new therapy clients, a \textbf{couple}, and the building started shaking like crazy. & a pair of people who live together  & couple (noun): two people who are married, engaged, or otherwise romantically involved with each other. \\
\midrule
\textbf{Worse} & Alimta is used together with cisplatin (another anticancer medicine) when the cancer is unresectable (cannot be removed by surgery alone) and malignant (has spread, or is likely to spread easily, to other parts of the body), in patients who have not received chemotherapy (medicines for cancer) before \textbf{advanced} or metastatic non-small cell lung cancer that is not affecting the squamous cells. & comparatively late in a course of development & advanced (adj.): Referring to a stage of disease, particularly cancer, that is more progressed, having spread significantly or becoming more serious in its development. \\
\midrule
\textbf{Wrong} & If they put a Republican into office, not only will they acquire less in terms of local \textbf{benefits} but their selected legislator will be relatively powerless to prevent other legislators from ''bringing home the bacon'' to their respective constituencies. & financial assistance in time of need & benefit (noun): An advantage or profit gained from something \\

\bottomrule
\end{tabular}
}
\caption{Examples of human annotation on the Definition Generation task.}
\label{tab:example-annotation-definition}
\end{table*} 

\begin{table*}[t]
\centering
\small
\resizebox{1\linewidth}{!}{
\begin{tabular}{p{1.7cm}p{5cm}p{2.5cm}p{5.8cm}} 
\toprule
\textbf{Annotation} & \textbf{Context} & \textbf{Gold Definition} & \textbf{GPT-4o Generation} \\
\midrule
\textbf{Good} & Ten shirt-sleeved ringers stand in a circle, one foot ahead of the other in a prize-fighter's stance, each pulling a \textbf{rope} that disappears through a small hole in the high ceiling of the ringing chamber. & a strong line  & In this sentence, the word "rope" refers to the thick, strong cord or line that each of the ten ringers is pulling on. These ropes are part of the mechanism used to ring bells, typically in a church or bell tower. The ropes extend upwards and disappear through a hole in the ceiling, suggesting they are connected to the bells above. The ringers pull on these ropes to create the movement needed to sound the bells, with each person handling one rope to control one bell. \\
\midrule
\textbf{Bad} & In this sense, a \textbf{teacher} can be compared to one's family doctor. & a person whose occupation is teaching & In the given sentence, the word "teacher" is being compared to a "family doctor", suggesting that a teacher, like a family doctor, plays a critical and foundational role in someone's life. Just as a family doctor provides care, guidance, and support for a person's health over time, a teacher provides knowledge, guidance, and support for a person's educational and personal development. This comparison highlights the importance and responsibility of a teacher in nurturing and shaping an individual's growth and well-being. \\
\midrule
\textbf{Borderline} & Haney hoped that nobody \textbf{noticed} his sudden pallor , as he felt the blood drain from his cheeks . & notice or perceive & In this sentence, "noticed" means that Haney hoped no one saw or became aware of his sudden pallor, or the change in color of his face, which became pale as he felt the blood draining from his cheeks. He did not want anyone to observe or pay attention to this change, possibly because it might reveal his nervousness or fear. \\
\bottomrule
\end{tabular}
}
\caption{Examples of human annotation on the Free-form Explanation task.}
\label{tab:example-annotation-free}
\end{table*} 

\begin{table*}[t]
\centering
\small
\resizebox{1\linewidth}{!}{
\begin{tabular}{p{1.7cm}p{5cm}p{2.5cm}p{5.8cm}} 
\toprule
\textbf{Annotation} & \textbf{Context} & \textbf{Gold Definition} & \textbf{GPT-4o Generation} \\
\midrule
\textbf{Yes} & Nobody witnessed the fall -- just the sickening impact when his body smashed on the pavement just outside the basement delivery \textbf{entrance}. & something that provides access (to get in or get out) & 1. The grand villa had an elaborate iron gate that served as the entrance to the sprawling estate.\newline
2. She waited nervously at the entrance of the conference hall, gathering her thoughts before heading inside to give her presentation. \newline
3. The amusement park's entrance was adorned with colorful lights and banners, inviting families to a day of fun and excitement.\\
\midrule
\textbf{Yes but same pattern} & One of the \textbf{reasons} American students do so poorly in math tests, as compared with British, French, German or Japanese students, is the influence of the ``New Math`` on American textbooks and teaching methods. & a justification for something existing or happening & 1. One of the reasons the restaurant consistently receives poor reviews is the inconsistent quality of its food and service.\newline
2. One of the reasons the city experiences frequent traffic congestion is the outdated infrastructure and lack of public transportation options.\newline
3. One of the reasons many people struggle to maintain a healthy lifestyle is the prevalence of fast food and sedentary work environments. \\
\midrule
\textbf{No} & So you \textbf{thought} I didn't mean what I said. & expect, believe, or suppose  & 1. You assumed I was joking, didn't you? \newline
2. You believed I wasn't serious about my statement? \newline
3. Did you suppose I was speaking in jest? \\
\midrule
\textbf{Borderline} & They may not be among the ''best and brightest '' of their \textbf{generation} -- there are very few such people, by definition. & all the people living at the same time or of approximately the same age & 1. The technological achievements of this generation will be studied by scholars for decades to come. \newline
2. Artists from her generation faced unique challenges but also had unprecedented opportunities to reach global audiences. \newline
3. Environmental concerns have become a defining feature of his generation, shaping policies and personal choices alike. \\
\bottomrule
\end{tabular}
}
\caption{Examples of human annotation on the Example Generation task.}
\label{tab:example-annotation-example}
\end{table*} 

\end{document}